\documentclass{article}

\usepackage{times}

\usepackage{graphicx} 
\usepackage{subfigure} 

\usepackage{natbib}

\usepackage{algorithm}
\usepackage{algorithmic}

\usepackage{hyperref}

\usepackage[accepted]{icml2013}

\usepackage{npbayes}

\icmltitlerunning{MAD-Bayes}

\begin{document} 

\twocolumn[
\icmltitle{MAD-Bayes: MAP-based Asymptotic Derivations from Bayes}

\icmlauthor{Tamara Broderick}{tab@stat.berkeley.edu}
\icmladdress{UC Berkeley, Statistics Department}
\icmlauthor{Brian Kulis}{kulis@cse.ohio-state.edu}
\icmladdress{Ohio State University, CSE Department}
\icmlauthor{Michael I. Jordan}{jordan@eecs.berkeley.edu}
\icmladdress{UC Berkeley, Statistics Department and EECS Department}

\icmlkeywords{K-means, clusters, features, unsupervised learning}

\vskip 0.3in
]

\begin{abstract} 
The classical mixture of Gaussians model is related to K-means via \emph{small-variance 
asymptotics}: as the covariances of the Gaussians tend to zero, the negative log-likelihood 
of the mixture of Gaussians model approaches the K-means objective, and the EM algorithm 
approaches the K-means algorithm.  \citet{kulis:2012:revisiting} used this observation
to obtain a novel K-means-like algorithm from a Gibbs sampler for the Dirichlet process
(DP) mixture.  We instead consider applying small-variance asymptotics directly to the 
posterior in Bayesian nonparametric models.  This framework is independent of any
specific Bayesian inference algorithm, and it has the major advantage that it
generalizes immediately to a range of models beyond the DP mixture.  To illustrate, 
we apply our framework to the feature learning setting, where the beta process and 
Indian buffet process provide an appropriate Bayesian nonparametric prior.  We obtain 
a novel objective function that goes beyond clustering to learn (and penalize new) 
groupings for which we relax the mutual exclusivity and exhaustivity assumptions of 
clustering.  We demonstrate several other algorithms, all of which are scalable and 
simple to implement.  Empirical results demonstrate the benefits of the new framework.
\end{abstract} 

\pagestyle{plain}
\setlength{\footskip}{0.4in}
\pagenumbering{arabic}

\section{Introduction} \label{sec:introduction}

Clustering is a canonical learning problem and arguably the dominant application of unsupervised learning.  Much of the popularity of clustering revolves around the K-means algorithm; its simplicity and scalability make it the preferred choice in many large-scale unsupervised learning problems---even though a wide variety of more flexible algorithms, including those from Bayesian nonparametrics, have been developed since the advent of K-means \citep{steinley:2006:k, jain:2010:data}. Indeed, \citet{berkhin:2006:survey} writes that K-means is
``by far the most popular clustering tool used nowadays in scientific and industrial applications.''

K-means does have several known drawbacks.  For one, the K-means algorithm clusters data into mutually exclusive and exhaustive clusters, which may not always be the optimal or desired form of latent structure for a data set.  For example, pictures on a photo-sharing website might each be described by multiple tags, or social network users might be described by multiple interests.  In these examples, a \emph{feature allocation} in which each data point can belong to any nonnegative integer number of groups---now called \emph{features}---is a more appropriate description of the data \citep{griffiths:2006:infinite,broderick:2013:feature}.  Second, the K-means algorithm requires advance knowledge of the number of clusters, which may be unknown in some applications.  
A vast literature exists just on how to choose the number of clusters using heuristics or extensions of K-means~\citep{steinley:2006:k,jain:2010:data}.
A recent algorithm called DP-means \citep{kulis:2012:revisiting} provides another perspective on the choice of cluster cardinality. Recalling the small-variance asymptotic argument that takes the EM algorithm for mixtures of Gaussians and yields the K-means algorithm, the authors apply this argument to a Gibbs sampler for a Dirichlet process (DP) mixture \citep{antoniak:1974:mixtures,escobar:1994:estimating,escobar:1995:bayesian} and obtain a K-means-like algorithm that does not fix the number of clusters upfront. 

Notably, the derivation 
of DP-means is specific to the choice of the sampling algorithm and is also not immediately 
amenable to the feature learning setting.  In this paper, we provide a more general perspective 
on these small-variance asymptotics.   We show that one can obtain the objective function for 
DP-means (independent of any algorithm) by applying asymptotics directly to the MAP estimation 
problem of a Gaussian mixture model with a Chinese Restaurant Process (CRP) prior \citep{blackwell:1973:ferguson,aldous:1985:exchangeability} on the 
cluster indicators.  The key is to express the estimation problem in terms of the exchangeable 
partition probability function (EPPF) of the CRP \citep{pitman:1995:exchangeable}.

A critical advantage of this more general view of small-variance asymptotics is that it provides a 
framework for extending beyond the DP mixture.  
The Bayesian nonparametric toolbox contains many models that may yield---via small-variance asymptotics---a range of new algorithms that to the best of our knowledge have not been
discovered in the K-means literature. We thus view our major contribution as providing new directions for researchers working on K-means and related discrete optimization problems in machine learning.

To highlight this generality, we show how
the framework may be used in the feature learning setting.  We take as our point of 
departure the beta process (BP) \citep{hjort:1990:nonparametric, ThibauxJo07}, which is the feature learning counterpart 
of the DP, and the Indian Buffet Process (IBP) \citep{griffiths:2006:infinite}, which is the feature learning counterpart of
the CRP.  We show how to express the corresponding MAP inference problem via an analogue
of the EPPF that we refer to as an ``exchangeable feature probability function'' (EFPF) \citep{broderick:2013:feature}.
Taking an asymptotic limit we obtain a novel objective function for feature learning, 
as well as a simple and scalable algorithm for learning features in a data set.  
The resulting algorithm, which we call \emph{BP-means}, is similar to the DP-means 
algorithm, but allows each data point to be assigned to more than one feature.  
We also use our framework to derive several additional algorithms, including algorithms
based on the Dirichlet-multinomial prior as well as extensions to the marginal MAP 
problem in which the cluster/feature means are integrated out.  We compare our algorithms 
to existing Gibbs sampling methods as well as existing hard clustering methods in order 
to highlight the benefits of our approach.

\section{MAP Asymptotics for Clustering} \label{sec:cluster}

We begin with the problem setting of \citet{kulis:2012:revisiting} 
but diverge in our treatment of the small-variance asymptotics.  We consider a Bayesian
nonparametric framework for generating data via a prior on clusterings and a likelihood 
that depends on the (random) clustering. Given prior and likelihood, we form a posterior 
distribution for the clustering. A point estimate of the clustering (i.e., a hard clustering) 
may be achieved by choosing a clustering that maximizes the posterior; the result is a 
\emph{maximum a posteriori} (MAP) estimate.

Consider a data set $x_{1}, \ldots, x_{N}$, where $x_{n}$ is a $D$-component vector. 
Let $\ko$ denote the (random) number of clusters. Let $z_{nk}$ equal one 
if data index $n$ belongs to cluster $k$ and 0 otherwise, so there is exactly one value of $k$ for each $n$ such that $z_{nk} = 1$. We can order the cluster labels $k$ so that the first $\ko$ are non-empty (i.e., $z_{nk} = 1$ for some $n$ for each such $k$). Together $\ko$ and $z_{1:N,1:\ko}$ describe a clustering.

The Chinese restaurant process (CRP) \citep{blackwell:1973:ferguson,aldous:1985:exchangeability} yields a prior on $\ko$ and $z_{1:N,1:\ko}$
as follows.  Let $\dpconc > 0$ be a hyperparameter of the model. The first customer 
(data index $1$) starts a new table in the restaurant; i.e., $z_{1,1} = 1$. 
Recursively, the $n$th customer (data index $n$) sits at an existing table $k$ 
with probability in proportion to the number of people sitting there (i.e., in 
proportion to $S_{n-1,k} := \sum_{m=1}^{n-1} z_{mk}$) and at a new table with 
probability proportional to $\dpconc$.

Suppose the final restaurant has $\ko$ tables with $N$ total customers sitting according to  $z_{1:N,1:\ko}$. Then the probability of this clustering is found by multiplying together the $N$ steps in the recursion described above:
\begin{equation}
	\label{eq:prior_crp}
	\mbp(z_{1:N,1:\ko}) = \dpconc^{\ko-1} \frac{\Gamma(\dpconc + 1)}{\Gamma(\dpconc + N)}
		\prod_{k=1}^{\ko} (S_{N,k} - 1)! ,
\end{equation}
a formula that is known as an exchangeable partition probability 
function (EPPF)~\cite{pitman:1995:exchangeable}.

As for the likelihood, a common choice is to assume that data in cluster $k$ are 
generated from a Gaussian with a cluster-specific mean $\mu_{k}$ and shared variance 
$\sigma^{2} I_{D}$ (where $I_{D}$ is the identity matrix of size $D \times D$ and 
$\sigma^{2} > 0$), and we will make that assumption here.  Suppose the $\mu_{k}$ 
are drawn iid Gaussian from a prior with mean 0 in every dimension and variance 
$\rho^{2} I_{D}$ for some hyperparameter $\rho^{2} > 0$: $\mbp(\mu_{1:\ko}) = \prod_{k=1}^{\ko} \gaus(\mu_{k} | 0, \rho^{2} I_{D})$. Then the likelihood of a data set $x = x_{1:N}$ given clustering $z = z_{1:N,1:\ko}$ and means $\mu = \mu_{1:\ko}$ is as follows:
$$
	\mbp(x | z, \mu) = \prod_{k=1}^{\ko} \prod_{n: z_{n,k}=1} \gaus(x_{n} | \mu_{k}, \sigma^{2} I_{D}).
$$

Finally, the posterior distribution over the clustering given the observed data, $\mbp(z, \mu | x)$, is calculated from the prior and likelihood using Bayes rule:
$
	\mbp(z, \mu | x) \propto \mbp(x | z, \mu) \mbp(\mu) \mbp(z).
$
We find the MAP point estimate for the clustering and cluster means by maximizing the posterior:
$ \argmax_{\ko,z,\mu} \mbp(z, \mu | x) $. Note that the point estimate will be the same if we instead minimize the negative log joint likelihood:
$ \argmin_{\ko,z,\mu} - \log \mbp(z, \mu, x) $.

In general, calculating the posterior or MAP estimate is difficult and usually requires 
approximation, e.g.\ via Markov Chain Monte Carlo or a variational method. A different 
approximation can be obtained by taking the limit of the objective function in this 
optimization as the cluster variances decrease to zero: $\sigma^{2} \rightarrow 0$. 
Since the prior allows an unbounded number of clusters, we expect that taking this 
limit alone will result in each data point being assigned to its own cluster. 
To arrive at a limiting objective function that favors a non-trivial cluster assignment, 
we modulate the number of clusters via the hyperparameter $\dpconc$, which varies linearly 
with the expected number of clusters in the prior. In particular, we choose some constant 
$\lambda^{2} > 0$ and let
$$
	\dpconc = \exp(-\lambda^{2} / (2 \sigma^{2})),
$$
so that, e.g., $\dpconc \rightarrow 0$ as $\sigma^{2} \rightarrow 0$.

Given this dependence of $\dpconc$ on $\sigma^{2}$ and letting $\sigma^{2} \rightarrow 0$, we find
that $- 2 \sigma^{2} \log \mbp(z, \mu, x)$ satisfies
\begin{equation}
	\label{eq:asymp_logp_crp}
  \sim \sum_{k=1}^{\ko} \sum_{n: z_{nk} = 1} \|x_{n} - \mu_{k} \|^{2} + (\ko - 1)\lambda^{2},
\end{equation}
 where $f(\sigma^{2}) \sim g(\sigma^{2})$ here denotes $f(\sigma^{2}) / g(\sigma^{2}) \rightarrow 1$ as $\sigma^{2} \rightarrow 0$. The double sum originates from the exponential function in the Gaussian data likelihood, and the penalty term---reminiscent of an AIC penalty \citep{akaike:1974:new}---originates from the CRP prior (\app{dp_obj}).
 
From \eq{asymp_logp_crp}, we see that finding the MAP estimate of the CRP Gaussian mixture model is asymptotically equivalent to the following optimization problem (\app{dp_obj}):
\begin{equation}
	\label{eq:crp_means_obj}
	\argmin_{\ko, z, \mu} \sum_{k=1}^{\ko} \sum_{n: z_{nk} = 1} \|x_{n} - \mu_{k} \|^{2} + (\ko - 1)\lambda^{2}.
\end{equation}
\citet{kulis:2012:revisiting} derived a similar objective function, which they called the {\em DP-means objective function} (a name we retain for \eq{crp_means_obj}), by first deriving a K-means-style algorithm from a DP Gibbs sampler. Here, by contrast, we have found this objective function directly from the MAP problem, with no reference to any particular inference algorithm and thereby demonstrating a more fundamental link between the MAP problem and \eq{crp_means_obj}.
In the following, we show that this focus on limits of a MAP estimate can yield useful optimization problems in diverse domains.

Notably, the objective in \eq{crp_means_obj} takes the form of the K-means objective function (the double sum) plus a penalty of $\lambda^{2}$ for each cluster after the first; this offset penalty is natural since any partition of a non-empty set must have at least one cluster.\footnote{The objective of \citet{kulis:2012:revisiting} penalizes all $\ko$ clusters; the optimal inputs are the same in each case.}
Once we have \eq{crp_means_obj}, we may consider efficient solution methods; one candidate is the DP-means algorithm of \citet{kulis:2012:revisiting}. 

\section{MAP Asymptotics for Feature Allocations} \label{sec:feature}

Once more consider a data set $x_{1:N}$, where $x_{n}$ is a $D$-component vector. Now let $\ko$ denote the (random) number of features in our model. Let $z_{nk}$ equal one if data index $n$ is associated with feature $k$ and zero otherwise; in the feature case, while there must be a finite number of $k$ values such that $z_{nk}=1$ for any value of $n$, it is not required (as in clustering) that there be exactly a single such $k$ or even any such $k$. We order the feature labels $k$ so that the first $\ko$ features are non-empty; i.e., we have $z_{nk}=1$ for some $n$ for each such $k$. Together $\ko$ and $z_{1:N,1:\ko}$ describe a feature allocation.

The Indian buffet process (IBP) \citep{griffiths:2006:infinite} is a prior on $z_{1:N,1:\ko}$ that places strictly 
positive probability on any finite, nonnegative value of $\ko$.  Like the CRP, 
it is based on an analogy between the customers in a restaurant and the data indices. 
In the IBP, the dishes in the buffet correspond to features. Let $\bpmass > 0$ be a 
hyperparameter of the model. The first customer (data index 1) samples $\ko_{1} \sim \pois(\bpmass)$ 
dishes from the buffet. Recursively, when the $n$th customer (data index $n$) arrives 
at the buffet, $\sum_{m=1}^{n-1} \ko_{m}$ dishes have been sampled by the previous 
customers. Suppose dish $k$ of these dishes has been sampled $S_{n-1,k}$ times by 
the first $n-1$ customers. The $n$th customer samples dish $k$ with probability 
$S_{n-1,k} / n$. The $n$th customer also samples $\ko_{n} \sim \pois(\bpmass/n)$ new dishes.

Suppose the buffet has been visited by $N$ customers who sampled a total of $\ko$ dishes. 
Let $z = z_{1:N,1:\ko}$ represent the resulting feature allocation. Let $H$ be the number 
of unique values of the $z_{1:N,k}$ vector across $k$; let $\tk_{h}$ be the number of $k$ 
with the $h$th unique value of this vector.  We calculate an ``exchangeable feature probability 
function'' (EFPF) \citep{broderick:2013:feature} by multiplying together the probabilities from the $N$ steps in the 
description and find that $\mbp(z)$ equals \citep{griffiths:2006:infinite}
\begin{align}
	\label{eq:prior_ibp}
	\frac{\bpmass^{\ko}
		\exp\left\{ -\sum_{n=1}^{N} \frac{\bpmass}{n} \right\}
		}{ \prod_{h=1}^{H} \tk_{h}! }
		\prod_{k=1}^{\ko} S_{N,k}^{-1} \binom{N}{S_{N,k}}^{-1}.
\end{align}

It remains to specify a probability for the observed data $x$ given the latent feature allocation $z$. The linear Gaussian model of \citet{griffiths:2006:infinite} is a natural extension of the Gaussian mixture model to the feature case. As previously, we specify a prior on feature means $\mu_{k} \stackrel{iid}{\sim} \gaus( 0, \rho^{2} I_{D})$ for some hyperparameter $\rho^{2} > 0$. Now data point $n$ is drawn independently with mean equal to the sum of its feature means, $\sum_{k=1}^{\ko} z_{nk} \mu_{k}$ and variance $\sigma^{2} I_{D}$ for some hyperparameter $\sigma^{2} > 0$. In the case where each data point belongs to exactly one feature, this model is just a Gaussian mixture. We often write the means as a $K \times D$ matrix $A$ whose $k$th row is $\mu_{k}$. Then, writing $Z$ for the matrix with $(n,k)$ element $z_{nk}$ and $X$ for the matrix with $n$th row $x_{n}$, we have $\mbp(X | Z, A)$ equal to
\begin{equation}
	\label{eq:linear_gaussian}
	\frac{1}{(2 \pi \sigma^{2})^{ND/2}} \exp\left\{ -\frac{ \tr((X - ZA)' (X - ZA)) }{2 \sigma^{2}} \right\}.
\end{equation}

As in the clustering case, we wish to find the joint MAP estimate of the structural component $Z$ and group-specific parameters $A$. It is equivalent to find the values of $Z$ and $A$ that minimize $- \log \mbp(X, Z, A)$. Finally, we wish to take the limit of this objective as $\sigma^{2} \rightarrow 0$. Lest every data point be assigned to its own separate feature, we modulate the number of features in the small-$\sigma^{2}$ limit by choosing some constant $\lambda^{2} > 0$ and setting
$
	\bpmass = \exp(- \lambda^{2} / (2\sigma^{2})).
$

Letting $\sigma^{2} \rightarrow 0$, we find that asymptotically (\app{bp_obj})
$$
-2\sigma^{2} \log \mbp(X, Z, A)
		\sim \tr[(X - ZA)' (X - ZA)] + \ko \lambda^{2},
$$
The trace originates from the matrix Gaussian, and the penalty term originates from the IBP prior.

It follows that finding the MAP estimate for the feature learning problem
is asymptotically equivalent to solving the following optimization problem:
\begin{equation}
	\label{eq:ibp_means_obj}
	\argmin_{\ko, Z, A} \tr[(X - ZA)' (X - ZA)] + \ko \lambda^{2}.
\end{equation}
We follow \citet{kulis:2012:revisiting} in referring to the underlying
random measure in denoting objective functions derived from Bayesian
nonparametric priors. Recalling that the beta process (BP)
\citep{hjort:1990:nonparametric, ThibauxJo07} is the 
random measure underlying the IBP, we refer to the objective function 
in \eq{ibp_means_obj} as the {\em BP-means objective}.  The trace term 
in this objective forms a K-means-style objective on a feature matrix 
$Z$ and feature means $A$ when the number of features (i.e., the number 
of columns of $Z$ or rows of $A$) is fixed. The second term enacts a penalty 
of $\lambda^{2}$ for each feature. In contrast to the DP-means objective, 
even the first feature is penalized since it is theoretically possible to 
have zero features.

\textbf{BP-means algorithm.}
We formulate a \emph{BP-means algorithm} to solve the optimization
problem in \eq{ibp_means_obj} and discuss its convergence properties.
In the following, note that $Z'Z$ is invertible so long as two features do not
have the same collection of indices. In this case, we simply combine
the two features into a single feature before performing the inversion.

\fbox{
\parbox{0.95 \columnwidth}{
\textbf{BP-means algorithm.}
Iterate the following two steps until no changes are made: \\
1. For $n = 1,\ldots,N$
	\begin{itemize}
		\item For $k = 1,\ldots,\ko$, choose the optimizing value (0 or 1) of $z_{nk}$.
		\item Let $Z'$ equal $Z$ but with one new feature (labeled $\ko+1$) containing only data index $n$.
		Set $A' = A$ but with one new row: $A'_{\ko+1,\cdot} \leftarrow X_{n,\cdot} - Z_{n,\cdot} A$.
		\item If the triplet $(\ko + 1, Z', A')$ lowers the objective from the triplet $(\ko, Z, A)$, replace the latter triplet with the former.
	\end{itemize}
2. Set $A \leftarrow (Z' Z)^{-1} Z' X$.
}
}

\begin{proposition} \label{prop:ibp_means}
The BP-means algorithm converges after a finite number of iterations to a local minimum of the BP-means objective in \eq{ibp_means_obj}.
\end{proposition}

See \app{bp_means_conv} for the proof.
Though the proposition guarantees convergence, it does not guarantee convergence to the global optimum---an analogous result to those available for the K-means and DP-means algorithms \citep{kulis:2012:revisiting}.
Many authors have noted the problem of local optima in the clustering literature \citep{steinley:2006:k,jain:2010:data}. One expects that the issue of local optima is only exacerbated in the feature domain, where the combinatorial landscape is much more complex. In clustering, this issue is often addressed by multiple random restarts and careful choice of cluster initialization; in \mysec{experiments} below, we also make use of random algorithm restarts and propose a feature initialization akin to one with provable guarantees for K-means clustering \citep{arthur:2007:k}.

\section{Extensions} \label{sec:extensions}

A number of variations on the Gaussian mixture posteriors are of 
interest in both nonparametric and parametric Bayesian inference. We briefly demonstrate
that our methodology applies readily to several such situations.

\subsection{Collapsed objectives} \label{sec:collapsed}

It is believed in many scenarios that {\em collapsing} out the
cluster or feature means from a Bayesian model by calculating instead the marginal 
structural posterior can improve MCMC sampler mixing~\citep{liu:1994:collapsed}.

\textbf{Clustering.} In the clustering case, collapsing 
translates to forming the posterior $\mbp(z | x) = \int_{\mu} \mbp(z, \mu | x)$.
Note that even in the cluster case, we may use the matrix representations $Z$, $X$, and
$A$ so long as we make the additional assumption that $\sum_{k=1}^{\ko} z_{nk} = 1$ for
each $n$. Finding the MAP
estimate $\argmax_{Z} \mbp(Z | X)$ may, as usual, be accomplished by minimizing the negative log joint distribution
with respect to $Z$. $\mbp(Z)$ is given by the CRP (\eq{prior_crp}). $\mbp(X|Z)$ takes the form
\citep{griffiths:2006:infinite}:
\begin{align}
	 	\label{eq:collapsed_data_like}
		\frac{ \exp\left\{ - \frac{ \tr\left( X' (I_{N} - Z (Z'Z + \frac{\sigma^{2}}{\rho^{2}} I_{D})^{-1} Z' ) X \right) }{2\sigma^{2}} \right\} }{ (2\pi \sigma^{2})^{(ND/2} (\rho^{2}/\sigma^{2})^{\ko D/2} |Z' Z + \frac{\sigma^{2}}{\rho^{2}} I_{D} |^{D/2}  }.
\end{align}
Using the same asymptotics in $\sigma^{2}$ and $\dpconc$ as before, we find the limiting optimization problem (\app{cdp_obj}):
\begin{equation}
	\label{eq:crp_collapse_obj_orig}
	 \argmin_{\ko, Z} \tr(X' (I_{N} - Z(Z'Z)^{-1} Z') X) + (\ko-1) \lambda^{2}.
\end{equation}
The first term in this objective was previously proposed, via independent considerations,
by \citet{gordon:1977:algorithm}.
Simple algebraic manipulations allow us to rewrite the objective in a
more intuitive format (\app{cdp_interp}):
\begin{equation}
	\label{eq:crp_collapse_obj}
	\argmin_{\ko, Z} \sum_{k=1}^{\ko} \sum_{n: z_{nk} = 1} \| x_{n,\cdot} - \bar{x}^{(k)} \|_{2}^{2} + (\ko-1) \lambda^{2},
\end{equation}
where $ \bar{x}^{(k)} := S_{N,k}^{-1} \sum_{m: z_{mk} = 1} x_{m,\cdot}$
is the $k$th empirical cluster mean, i.e.\ the mean of all data points
assigned to cluster $k$.
This {\em collapsed DP-means objective} is just the original
DP-means objective in \eq{crp_means_obj} with the cluster means replaced
by empirical cluster means.

A corresponding optimization algorithm is as follows.
\fbox{
\parbox{0.95 \columnwidth}{
\textbf{Collapsed DP-means algorithm.}
Repeat the following step until no changes are made: \\
1. For $n = 1,\ldots,N$
	\begin{itemize}
		\item Assign $x_{n}$ to the closest cluster if the contribution to the objective in \eq{crp_collapse_obj} from the squared distance is at most $\lambda^{2}$.
		\item Otherwise, form a new cluster with just $x_{n}$.
	\end{itemize}
}
}
A similar proof to that of \citet{kulis:2012:revisiting} shows that this algorithm
converges in a finite number of iterations to a local minimum
of the objective.

\textbf{Feature allocation.} We have already noted
that the likelihood associated with the Gaussian mixture
model conditioned on a clustering is just a special case of
the linear Gaussian model conditioned on a feature matrix.
Therefore, it is not surprising that \eq{collapsed_data_like}
also describes $\mbp(X|Z)$ when $Z$ is a feature matrix.
Now, $\mbp(Z)$ is given by the IBP (\eq{prior_ibp}). 
Using the same asymptotics in
$\sigma^{2}$ and $\bpmass$ as in 
the joint MAP case, the MAP problem for feature allocation
$Z$ asymptotically becomes (\app{cbp_obj}):
\begin{equation}
	\label{eq:ibp_collapse_obj}
	 \argmin_{\ko, Z} \tr(X' (I_{N} - Z(Z'Z)^{-1} Z') X) + \ko \lambda^{2}.
\end{equation}
The key difference with \eq{crp_collapse_obj_orig} is that here $Z$ may have any finite
number of ones in each row. We call the objective in \eq{ibp_collapse_obj} the {\em
collapsed BP-means objective}.

Just as the collapsed DP-means objective had an empirical cluster means
interpretation, so does the collapsed BP-means objective have an interpretation
in which the feature means matrix $A$ in the BP-means objective (\eq{ibp_means_obj})
is replaced by its empirical
estimate $(Z'Z)^{-1}Z$.
In particular, we can rewrite the objective in \eq{ibp_collapse_obj} as
$$
	\tr[(X - Z(Z'Z)^{-1}Z'X)' (X - Z(Z'Z)^{-1}Z'X)] + \ko \lambda^{2}.
$$
A corresponding optimization algorithm is as follows.
\fbox{
\parbox{0.95 \columnwidth}{
\textbf{Collapsed BP-means algorithm.}
Repeat the following step until no changes are made: \\
1. For $n = 1,\ldots,N$
	\begin{itemize}
		\item Choose $z_{n,1:\ko}$ to minimize the objective in \eq{ibp_collapse_obj}. Delete any redundant features.
		\item Add a new feature (indexed $\ko+1$) with only data index $n$ if doing so decreases the objective and if the feature would not be redundant.
	\end{itemize}
}
}
A similar proof to that of \prop{ibp_means} shows that this algorithm
converges in a finite number of iterations to a local minimum
of the objective.

\subsection{Parametric objectives} \label{sec:finite}

The generative models studied so far are {\em nonparametric} in the usual Bayesian
sense; there is no a priori bound on the number of cluster or feature
parameters. The objectives
above are similarly nonparametric. Parametric models,
with a fixed bound on the number of clusters or features, are often useful as well,
and we explore these here.

First, consider a clustering prior with some fixed
maximum number of clusters $K$. Let $q_{1:K}$ represent a distribution over
clusters. Suppose $q_{1:K}$ is drawn from a finite Dirichlet distribution with
size $K > 1$ and parameter $\dpconc > 0$. Further, suppose the cluster
for each data point is drawn iid according to $q_{1:K}$. Then, integrating out $q$,
the marginal distribution of the clustering is Dirichlet-multinomial:
\begin{equation}
	\label{eq:cluster_finite_prior}
	\mbp(z) = \frac{\Gamma(K \dpconc)}{\Gamma(N + K \dpconc)}
		\prod_{k=1}^{K} \frac{\Gamma(S_{N,k} + \dpconc)}{\Gamma(\dpconc)}.
\end{equation}
We again assume a Gaussian mixture likelihood, only now the number
of cluster means $\mu_{k}$ has an upper bound of $K$.

We can find the MAP estimate of $z$ and $\mu$ under this model
in the limit $\sigma^{2} \rightarrow 0$.
With $\dpconc$ fixed, the clustering prior has no effect, and the resulting
optimization problem is
$
	\argmin_{z, \mu} \sum_{k=1}^{K} \sum_{n: z_{nk} = 1} \|x_{n} - \mu_{k} \|^{2},
$
which is just the usual K-means optimization problem.

We can also try scaling
$\dpconc = \exp(-\lambda^{2}/(2\sigma^{2}))$ for some constant
$\lambda^{2} > 0$ as in the unbounded cardinality case.
Then taking the $\sigma^{2} \rightarrow 0$ limit of the log joint likelihood
yields a term of $\lambda^{2}$ for
each cluster containing at least one data index in the product in
\eq{cluster_finite_prior}---except
for one such cluster. Call
the number of such activated clusters $\ko$. The resulting optimization problem is
\begin{equation}
	\label{eq:finite_cluster_obj}
	\argmin_{\ko, z, \mu} \sum_{k=1}^{K} \sum_{n: z_{nk} = 1} \|x_{n} - \mu_{k} \|^{2} + (K \wedge \ko - 1) \lambda^{2}.
\end{equation}
This objective caps the number of clusters at $K$ but contains a penalty for each new
cluster up to $K$.

A similar story holds in the feature case. Imagine that we have a fixed maximum of $K$ features. In this finite case, we now let $q_{1:K}$ represent frequencies of each feature and let $q_{k} \stackrel{iid}{\sim} \tb(\bpmass,1)$. We draw $z_{nk} \sim \bern(q_{k})$ iid across $n$ and independently across $k$. The linear Gaussian likelihood model is as in \eq{linear_gaussian} except that now the number of features is bounded. If we integrate out the $q_{1:K}$, the resulting marginal prior on $Z$ is
\begin{equation}
	\label{eq:finite_feature_prior}
	\prod_{k=1}^{K} \left(
		\frac{\Gamma(S_{N,k} + \bpmass) \Gamma(N - S_{N,k} + 1)}{\Gamma(N + \bpmass + 1) }
		\frac{\Gamma(\bpmass + 1)}{\Gamma(\bpmass)\Gamma(1)} \right).
\end{equation}
Then the limiting MAP
problem as $\sigma^{2} \rightarrow 0$ is
\begin{equation}
	\label{eq:feature_k_means_obj}
	\argmin_{Z, A} \tr[(X - ZA)' (X - ZA)].
\end{equation}
This objective is analogous to the K-means objective but holds for the more general
problem of feature allocations.

\eq{feature_k_means_obj} can be solved according to a K-means-style
algorithm. Notably, in the following algorithm, all of the optimizations for $n$ in step 1
may be performed in parallel.
\fbox{
\parbox{0.95 \columnwidth}{
\textbf{K-features algorithm.}
Repeat the following steps until no changes are made: \\
1. For $n = 1,\ldots,N$
	\begin{itemize}
		\item For $k=1,\ldots,K$, set $z_{n,k}$ to minimize $\| x_{n,1:K} - z_{n,1:K} A \|^{2}$.
	\end{itemize}
2. Set $A = (Z'Z)^{-1}Z' X$.
}
}

We can further set $\bpmass = \exp(-\lambda^{2}/(2\sigma^{2}))$ as
for the unbounded cardinality case before taking the limit $\sigma^{2} \rightarrow 0$.
Then a $\lambda^{2}$ term contributes to the limiting objective for
each non-empty feature from the product in \eq{finite_feature_prior}. The resulting objective is
\begin{equation}
	\label{eq:finite_feat_obj}
	\argmin_{\ko, Z, A} \tr[(X - ZA)' (X - ZA)] + (K \wedge \ko) \lambda^{2},
\end{equation}
reminiscent of the BP-means objective but with a cap of $K$ possible features.

\section{Experiments} \label{sec:experiments}

We examine collections of unlabeled data
to discover latent shared features. We have already seen
the BP-means and collapsed BP-means algorithms for
learning these features when the number of features
is unknown. A third algorithm that we evaluate here
involves running the K-features algorithm for different 
values of $K$ and choosing the joint values of $K, Z, A$ 
that minimize the BP-means objective in \eq{ibp_means_obj};
we call this the \emph{stepwise K-features algorithm}. 
If we assume the plot of the minimized K-features objective
(\eq{feature_k_means_obj})
as a function of $K$ has increasing increments, then we need 
only run the K-features algorithm for increasing $K$ until the objective 
decreases.

It is well known that the K-means algorithm is sensitive to the
choice of cluster initialization \citep{pena:1999:empirical}. Potential methods of addressing
this issue include multiple random initializations and choosing
initial, random cluster centers according to the K-means++ algorithm~\citep{arthur:2007:k}.
In the style of K-means++, we introduce a similar feature means
initialization.

We first consider fixed $K$. In K-means++, the
initial cluster center is chosen uniformly at random from the 
data set. However, we note that empirically, the various 
feature algorithms discussed tend to prefer the creation of
a {\em base feature}, shared amongst all the data.
So start by assigning every data index to the first feature, and let the first
feature mean be the mean of all the data points. 
Recursively, for feature $k$ with $k > 1$,
calculate the distance from each data point $x_{n,\cdot}$
to its feature representation $z_{n,\cdot} A$ for the construction
thus far. Choose a data index $n$ with probability proportional to this
distance squared. Assign $A_{k,\cdot}$ to be the $n$th distance.
Assign $z_{m,k}$ for all $m = 1,\ldots,N$ to optimize the K-features
objective.
In the case where $K$ is not known in advance, we repeat the recursive
step as long as doing so decreases the objective.

Another important consideration in running these algorithms without
a fixed number of clusters or features is choosing the relative penalty effect
$\lambda^{2}$. One option is to solve for $\lambda^{2}$ from a proposed $K$
value via a heuristic \citep{kulis:2012:revisiting} or validation on a data subset.
Rather than assume $K$ and return to it in this roundabout way, in the following
we aim merely to demonstrate that
there exist reasonable values of $\lambda^{2}$ that return meaningful results.
More carefully examining the translation from a discrete ($K$) to continuous
($\lambda^{2}$) parameter
space may be a promising direction for future work.

\subsection{Tabletop data} \label{sec:table}

Using a LogiTech digital webcam,
\citet{griffiths:2006:infinite} took 100 pictures of four objects
(a prehistoric handaxe, a Klein bottle, a cellular phone, and a
\$20 bill)
placed on a tabletop. The images are in JPEG format with 240 pixel
height, 320 pixel width, and 3 color channels. Each object
may or may not appear in a given picture; the experimenters
endeavored to place each object (by hand) in a respective
fixed location across pictures.

This setup lends itself naturally to the feature allocation
domain. We expect to find a base feature
depicting the tabletop and four more features, respectively corresponding to
each of the four distinct objects. Conversely, clustering on this
data set would yield
either a cluster for each distinct feature combination---a much
less parsimonious and less informative representation
than the feature allocation---or some averages over feature combinations. The latter
case again fails to capture the combinatorial nature of the data.

We emphasize a further point about identifiability within this combinatorial structure.
One ``true'' feature allocation for this data is the one described above.
But an equally valid allocation, from a combinatorial perspective,
is one in which the base feature contains all four objects and the tabletop.
Then there are four further features, each of which deletes an object
and replaces it with tabletop; this allows every possible combination
of objects on the tabletop to be constructed from the features.
Indeed, any combination of objects on the tabletop
could equally well serve as a base feature; the four remaining features
serve to add or delete objects as necessary.

We run PCA on the data and keep the first $D=100$ principal components
to form the data vector for each image. This pre-processing is the same
as that performed by \citet{griffiths:2006:infinite}, except the authors
in that case first average the three color channels of the images.

\begin{figure}
\begin{minipage}{.67\columnwidth}
\begin{center}
\begin{tabular}{l | llll}
Alg & Per run & Total & \#\\ \hline
Gibbs & $8.5 \cdot 10^{3}$ & --- & 10 \\
Collap & 11 & $1.1 \cdot 10^{4}$ & 5 \\
BP-m & 0.36 & $3.6 \cdot 10^{2}$ & 6 \\
FeatK & 0.10 & $1.55 \cdot 10^{2}$ & 5
\end{tabular}
\end{center}
\end{minipage}
\hfill
\begin{minipage}{.31\columnwidth}
\begin{center}
\includegraphics[width=0.99\columnwidth]{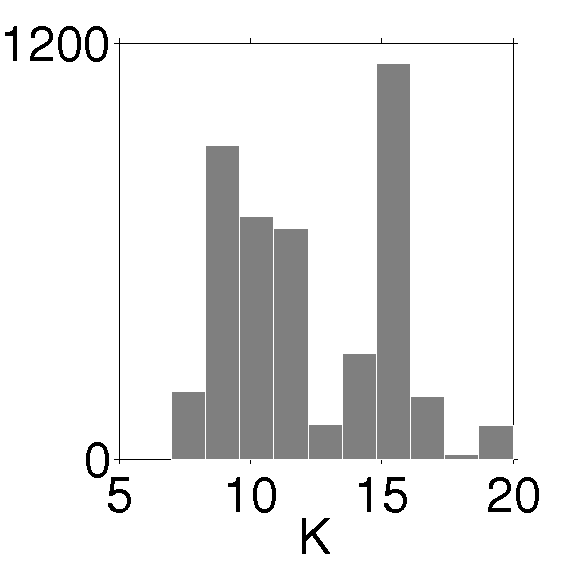}
\end{center}
\end{minipage}
\caption{\label{fig:comp_tabletop}
\emph{Left}: A comparison of results for the
IBP Gibbs sampler \citep{griffiths:2006:infinite}, the collapsed BP-means algorithm, 
the basic BP-means algorithm, and the stepwise K-features algorithm.
The first column shows the time for each run of the algorithm in seconds; the second 
column shows the total running time of the algorithm (i.e., over multiple 
repeated runs for the final three); and the third column shows the final 
number of features learned (the IBP \# is stable for $>900$ final iterations). \emph{Right}: A histogram of collections of the final
$K$ values found by the
IBP for a variety of initializations and parameter starting values.
}
\end{figure}

We consider the Gibbs sampling algorithm of \citet{griffiths:2006:infinite}
with initialization (mass parameter $1$ and feature mean variance $0.5$) and number of sampling steps (1000) determined by the authors;
we explore alternative initializations below. We compare
to the three feature means algorithms described above---all with $\lambda^{2}=1$.
Each of the final three algorithms uses the appropriate
variant of greedy initialization analogous to K-means++.
We run 1000 random initializations of the collapsed and BP-means algorithms
to mitigate issues of local minima. We run 300 random initializations of the
K-features algorithm for each value of $K$ and
note that $K=2,\ldots,6$ are (dynamically) explored by
the algorithm. All code was run in Matlab on the same computer.
Timing and feature count results are shown on the left of \fig{comp_tabletop}.

\begin{figure}
\begin{tabular}{@{}c@{\hspace{0.1cm}}c@{\hspace{0.1cm}}c@{\hspace{0.1cm}}c}
\includegraphics[width=0.24\columnwidth]{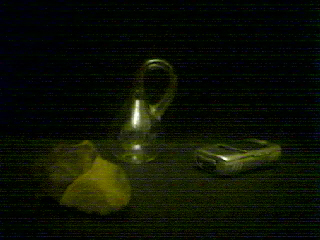}
& \includegraphics[width=0.24\columnwidth]{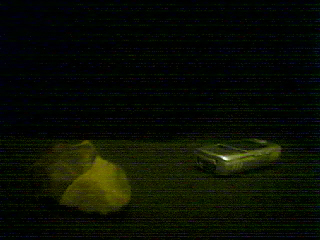}
& \includegraphics[width=0.24\columnwidth]{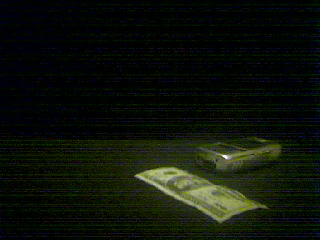}
& \includegraphics[width=0.24\columnwidth]{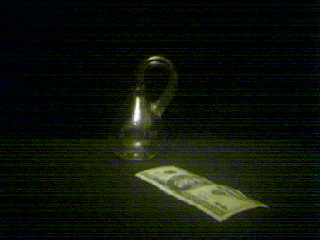} \\
10111
& 11111
& 11010
& 10000 \\
\includegraphics[width=0.24\columnwidth]{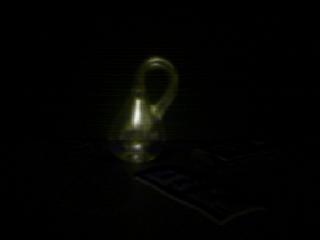}
& \includegraphics[width=0.24\columnwidth]{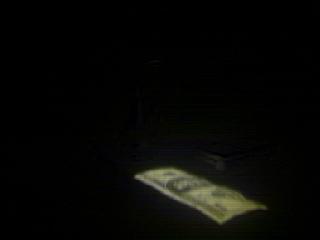}
& \includegraphics[width=0.24\columnwidth]{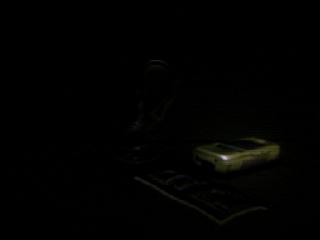}
& \includegraphics[width=0.24\columnwidth]{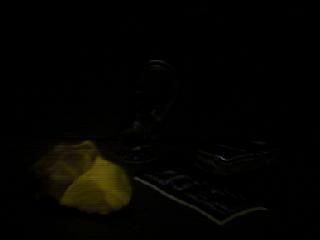} \\
(subtract) 
& (subtract)
& (add)
& (add)
\end{tabular}
\caption{ \label{fig:tabletop}
{\em Upper row}: Four example images in the tabletop
data set. {\em Second row}: Feature assignments of each
image. The first feature is the base feature, which
depicts the Klein bottle and \$20 bill on a tabletop and is almost
identical to the fourth picture in the first row. The remaining four
features are shown in order in the {\em third row}. The {\em fourth row}
indicates whether the picture is added or subtracted when the feature is present.
}
\end{figure}

While it is notoriously difficult to compare computation times for
deterministic, hard-assignment algorithms such K-means to stochastic 
algorithms such as Gibbs sampling, particularly given the practical
need for reinitialization to avoid local minima in the former, and difficult-to-assess
convergence in the latter, it should be clear from the first column in the lefthand table
of \fig{comp_tabletop} that there is a major difference
in computation time between Gibbs sampling and the new algorithms.  
Indeed, even when the BP-means algorithm is run 1000 times in a reinitialization 
procedure, the total time consumed by the algorithm is still an order
of magnitude less than that for a single run of Gibbs sampling.
We note also that stepwise K-features 
is the fastest of the new algorithms.

We further note that if we were to take advantage of parallelism,
additional drastic advantages could be obtained for the new algorithms.
The Gibbs sampler requires each Gibbs iteration to be performed sequentially 
whereas the random initializations of the various feature means algorithms 
can be performed in parallel.  A certain level of parallelism may even be
exploited for the steps within each iteration
of the collapsed and BP-means algorithms while
the $z_{n,1:K}$ optimizations of repeated feature K-means may 
all be performed 
in parallel across $n$ (as in classic K-means).

Another difficulty in comparing algorithms is that there is no clear single
criterion with which to measure accuracy of the final model in unsupervised
learning problems such as these. We do note, however, that
theoretical considerations suggest that the IBP is not designed
to find either a fixed number of features as $N$ varies nor roughly equal sizes in those
features it does find~\citep{broderick:2012:beta}. This observation
may help explain the distribution of observed feature counts over
a variety of IBP runs with the given data. To obtain feature counts from the IBP,
we tried running in a variety of different scenarios---combining different
initializations (one shared feature, 5 random features, 10 random features, initialization
with the BP-means result) and different starting parameter values\footnote{We found convergence failed for some parameter initializations 
outside this range} (mass parameter
values ranging logarithmically from 0.01 to 1 and mean-noise parameter values
ranging logarithmically from
0.1 to 10).
The final hundred $K$ draws for each of these combinations
are combined and summarized in a histogram on the right
of \fig{comp_tabletop}. Feature counts lower than 7 were not obtained in our 
experiments, which suggests these values are, at least, difficult to obtain using
the IBP with the given hyperpriors.

On the other hand,
the feature counts for the new K-means-style algorithms suggest
parsimony is more easily achieved in this case.
The lower picture and text rows of \fig{tabletop} show the features 
(after the base feature) found by feature K-means: as desired, there is 
one feature per tabletop object. The upper text row of \fig{tabletop} 
shows the features to which each of the example images in the top row are 
assigned by the optimal feature allocation. For comparison, the collapsed 
algorithm also finds an optimal feature encoding. The BP-means algorithm
adds an extra, superfluous feature containing both the Klein bottle and \$20 bill.

\subsection{Faces data} \label{sec:faces}

Next, we analyze the FEI face database, consisting of 400 pictures of pre-aligned
faces \citep{thomaz:2010:new}.
200 different individuals are pictured, each with one smiling
and one neutral expression. Each picture has height 300 pixels,
width 250 pixels, and one grayscale channel. Four example
pictures appear in the first row of \fig{face}. This time, we compare
the repeated feature K-means algorithm to classic K-means.
We keep the top 100 principal components
to form the data vectors for both algorithms.

\newlength{\tabseploc}
\setlength{\tabseploc}{0.1cm}
\newlength{\tabcolloc}
\setlength{\tabcolloc}{0.16\columnwidth}
\newlength{\picsize}
\setlength{\picsize}{\tabcolloc}
\begin{figure}
\parskip=0cm
\begin{center}
\begin{tabular}{c@{\hspace{\tabseploc}}p{\picsize}@{\hspace{\tabseploc}}p{\picsize}@{\hspace{\tabseploc}}p{\picsize}@{\hspace{\tabseploc}}p{\picsize} p{1.5\picsize}}
\begin{sideways} Samples \end{sideways}
& \includegraphics[width=\tabcolloc]{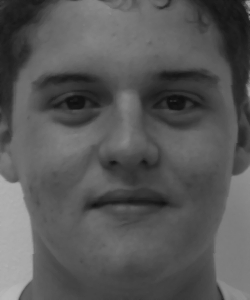}
& \includegraphics[width=\tabcolloc]{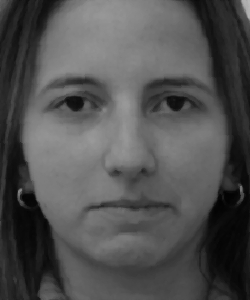}
& \includegraphics[width=\tabcolloc]{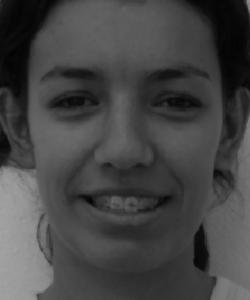}
& \includegraphics[width=\tabcolloc]{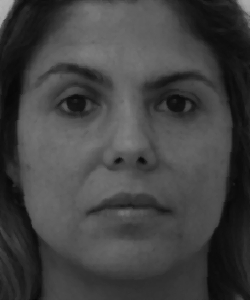}
& \\
\begin{sideways} 3 features \end{sideways}
& \includegraphics[width=\tabcolloc]{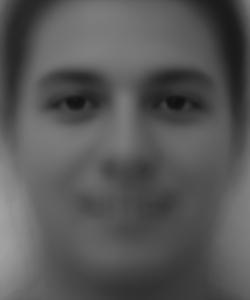}
& \includegraphics[width=\tabcolloc]{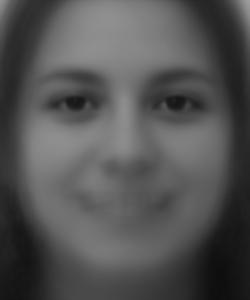}
& \includegraphics[width=\tabcolloc]{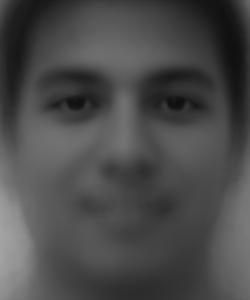}
& \multicolumn{2}{c}{\hspace{-1cm}\raisebox{1.1cm}{\parbox{2\picsize}{3 feature assign: \\
											100,110,101,111}}} \\
\begin{sideways} 3 clusters \end{sideways}
& \includegraphics[width=\tabcolloc]{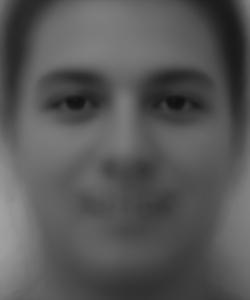}
& \includegraphics[width=\tabcolloc]{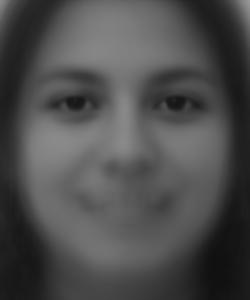}
& \includegraphics[width=\tabcolloc]{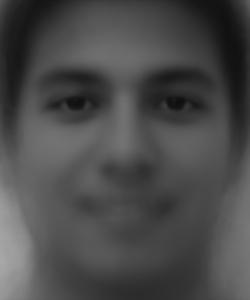}
& \multicolumn{2}{c}{\hspace{-1cm}\raisebox{0.7cm}{\parbox{2\picsize}{3 cluster assign: \\
											1,2,3,2 \\
										4 cluster assign: \\
											1,2,3,2}}} \\
\begin{sideways} 4 clusters \end{sideways}
& \includegraphics[width=\tabcolloc]{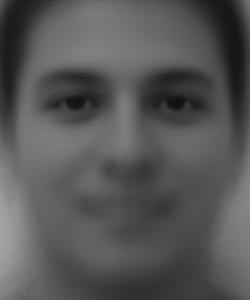}
& \includegraphics[width=\tabcolloc]{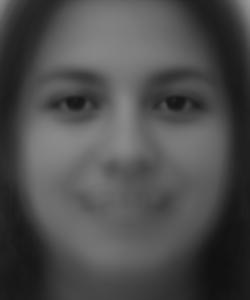}
& \includegraphics[width=\tabcolloc]{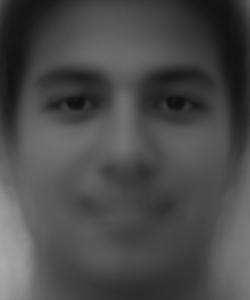}
& \includegraphics[width=\tabcolloc]{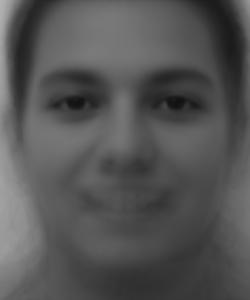}
& \\
\end{tabular}
\end{center}
\caption{ \label{fig:face}
{\em First row}: Four sample faces. {\em Second row}: The base feature (left) and other 2 features returned by repeated feature K-means
with $\lambda^{2} = 5$. The final pictures are the cluster means from K-means with $K=3$ ({\em third row}) and $K=4$ ({\em fourth row}). The righthand text shows how the sample pictures (left to right) are assigned to features and clusters by each algorithm.
}
\end{figure}

With a choice of $\lambda^{2} = 5$, repeated feature K-means
chooses one base feature (lefthand picture in the second row
of \fig{face}) plus two additional features as optimal; the central and righthand pictures
in the second row of \fig{face} depict the sum of the base feature plus
the corresponding feature. The base feature is
a generic face. The second feature codes for longer hair and a shorter chin.
The third feature codes for darker skin and slightly different
facial features. The feature combinations of each picture in the first row
appear in the first text row on the right; all four possible combinations are
represented. 

K-means with 2 clusters and feature K-means with 2 features
both encode exactly 2 distinct, disjoint groups. For larger numbers of groups though, the
two representations diverge. For instance, consider a 3-cluster model of the face data, which has the same number of parameters as the 3-feature model.
The resulting cluster means appear in the third row
of \fig{face}. While the cluster means appear similar to the feature means,
the assignment of faces to clusters is quite different. The second righthand text row in 
\fig{face} shows to which cluster each of the four first-row faces is assigned.
The feature allocation of the fourth picture in the top row tells us that
the subject has long hair and certain facial features, roughly, whereas the clustering
tells us that the subject's hair is more dominant than facial structure in determining grouping. 
Globally, the counts of faces for clusters (1,2,3) are (154,151,95) while
the counts of faces for feature combinations (100,110,101,111) are (139,106,80,75).

We might also consider a clustering of size 4 since there are 4 groups specified
by the 3-feature model. The resulting cluster means are in the bottom row of
\fig{face}, and the cluster assignments of the sample pictures are in the bottom, righthand text row.
None of the sample pictures falls in cluster 4. Again, the groupings
provided by the feature allocation and the clustering are quite different. Notably, the 
clustering has divided up the pictures with shorter hair into 3 separate clusters.
In this case, the counts of faces for clusters (1,2,3,4) are (121,150,74,55).
The feature allocation here seems to provide a sparser representation and more
interpretable groupings relative to both cluster cardinalities.

\section{Conclusions} \label{sec:concl}

We have developed a general methodology for obtaining hard-assignment
objective functions from Bayesian MAP problems. The key idea is to include
the structural variables explicitly in the posterior using combinatorial
functions such as the EPPF and the EFPF.  We apply this methodology to a 
number of generative models for unsupervised learning, with particular
emphasis on latent feature models. We show that the resulting algorithms
are capable of modeling latent structure out of reach of clustering algorithms
but are also much faster than existing feature allocation learners studied
in Bayesian nonparametrics. 
We have devoted some effort to finding 
algorithmic optimizations in the style of K-means (e.g., extending K-means++
initializations) in this 
domain. Nonetheless, the large literature on optimal initialization and fast, distributed
running of the K-means algorithm suggests that, with some thought, the
algorithms presented here can still be much improved in future work.

\section{Acknowledgments} \label{sec:ack}

We would like to thank Tom Griffiths for generously sharing his code.
TB's research was supported by a National Science Foundation
Graduate Research Fellowship and a Berkeley Fellowship. BK's 
research was supported by NSF award IIS-1217433.
This material is based upon work supported in part by the Office of
Naval Research under contract/grant number N00014-11-1-0688.

\bibliography{kmeans}
\bibliographystyle{icml2013}

\newpage

\appendix
\allowdisplaybreaks
\section*{Supplementary Material}

\section{DP-means objective derivation} \label{app:dp_obj}

First consider the generative model in \mysec{cluster}. 
The joint distribution of the observed data $x$, cluster indicators $z$, and cluster means $\mu$ can be written as follows.
\begin{align*}
	\mbp(x,z,\mu) &= \mbp(x|z,\mu) \mbp(z) \mbp(\mu) \\
		&= 
		\prod_{k=1}^{\ko} \prod_{n: z_{n,k}=1} \gaus(x_{n} | \mu_{k}, \sigma^{2} I_{D}) \\
		& {} \cdot \dpconc^{\ko-1} \frac{\Gamma(\dpconc + 1)}{\Gamma(\dpconc + N)} \prod_{k=1}^{\ko} (S_{N,k} - 1)! \\
		& {} \cdot \prod_{k=1}^{\ko} \gaus(\mu_{k} | 0, \rho^{2} I_{D})
\end{align*}

Then set $\theta := \exp(-\lambda^{2}/(2\sigma^{2}))$ and consider the limit $\sigma^{2} \rightarrow 0$. In the following, $f(\sigma^{2}) = O(g(\sigma^{2}))$ denotes that there exist some constants $c, s^{2} > 0$ such that $|f(\sigma^{2})| \le c |g(\sigma^{2})|$ for all $\sigma^{2} < s^{2}$.
\begin{align*}
	\lefteqn{ -\log \mbp(x,z,\mu) } \\
		&=
		\sum_{k=1}^{\ko} \sum_{n: z_{n,k}=1} \left[ 
			O(\log \sigma^{2})
			+ \frac{1}{2 \sigma^{2}} \| x_{n} - \mu_{k} \|^{2}
			\right] \\
		& {} + (\ko - 1) \frac{ \lambda^{2} }{ 2 \sigma^{2} }
			+ O(1)
			\\
		& {} + O(1)
\end{align*}

It follows that
\begin{align*}
	-2\sigma^{2} \log \mbp(x,z,\mu)
		&= \sum_{k=1}^{\ko} \sum_{n: z_{n,k}=1} \| x_{n} - \mu_{k} \|^{2} \\
		& {} + (\ko - 1) \lambda^{2} + O(\sigma^{2} \log(\sigma^{2})).
\end{align*}
But since $\sigma^{2} \log(\sigma^{2}) \rightarrow 0$ as $\sigma^{2} \rightarrow 0$, we have that the remainder of the righthand side is asymptotically equivalent (as $\sigma^{2} \rightarrow 0$) to the lefthand side (\eq{asymp_logp_crp}).

\section{BP-means objective derivation} \label{app:bp_obj}

The recipe is the same as in \app{dp_obj}. This time we start with the 
generative model in \mysec{feature}. The joint distribution of the observed data $X$,
feature indicators $Z$, and feature means $A$ can be written as follows.
\begin{align*}
	\lefteqn{ \mbp(X,Z,A) = \mbp(X|Z,A) \mbp(Z) \mbp(A) } \\
		&=
		\frac{1}{(2 \pi \sigma^{2})^{ND/2}} \exp\left\{ -\frac{1}{2\sigma^{2}} \tr((X - ZA)' (X - ZA)) \right\} \\
		& {} \cdot \frac{\bpmass^{\ko}
			\exp\left\{ -\sum_{n=1}^{N} \frac{\bpmass}{n} \right\}
			}{ \prod_{h=1}^{H} \tk_{h}! }
			\prod_{k=1}^{\ko} \frac{(S_{N,k}-1)! (N - S_{N,k})!}{N!} \\
		& {} \cdot 
		\frac{1}{(2 \pi \rho^{2})^{\ko D/2}} \exp\left\{ -\frac{1}{2 \rho^{2}} A' A \right\}
\end{align*}
		
Now set $\gamma := \exp(-\lambda^{2}/(2\sigma^{2}))$ and consider the limit $\sigma^{2} \rightarrow 0$.
Then
\begin{align*}
	\lefteqn{ -\log \mbp(X,Z,A) } \\
		&=
		O(\log \sigma^{2})
			+ \frac{1}{2\sigma^{2}} \tr((X - ZA)' (X - ZA)) \\
		& {} + \ko \frac{ \lambda^{2} }{ 2 \sigma^{2} }
			+ \exp(-\lambda^{2}/(2\sigma^{2})) \sum_{n=1}^{N} n^{-1}
			+ O(1) \\
		& {} + O(1)
\end{align*}

It follows that
\begin{align*}
	-2\sigma^{2} &\log \mbp(X,Z,A) 
		= \tr[(X - ZA)' (X - ZA)] + \ko \lambda^{2} \\
		& {} + O\left(\sigma^{2} \exp(-\lambda^{2}/(2\sigma^{2}))\right) + O(\sigma^{2} \log(\sigma^{2})).
\end{align*}
But since $\exp(-\lambda^{2}/(2\sigma^{2})) \rightarrow 0$ and $\sigma^{2} \log(\sigma^{2}) \rightarrow 0$ as $\sigma^{2} \rightarrow 0$, we have that $-2\sigma^{2} \log  \mbp(X,Z,A) \sim \tr[(X - ZA)' (X - ZA)] + \ko \lambda^{2}$.

\section{Collapsed DP-means objective derivation} \label{app:cdp_obj}

We apply the usual recipe as in \app{dp_obj}. The generative model for
collapsed DP-means is described in \mysec{collapsed}. The joint distribution
of the observed data $X$ and cluster indicators $Z$ can be written as follows.
\begin{align*}
	\lefteqn{ \mbp(X,Z) = \mbp(X|Z) \mbp(Z) } \\
		&= 
		\left( (2\pi)^{ND/2} (\sigma^{2})^{(N-\ko )D/2} (\rho^{2})^{\ko D/2} |Z' Z + \frac{\sigma^{2}}{\rho^{2}} I_{D} |^{D/2} \right)^{-1} \\
		& {} \cdot \exp\left\{ - \frac{1}{2\sigma^{2}} \tr\left( X' (I_{N} - Z (Z'Z + \frac{\sigma^{2}}{\rho^{2}} I_{D})^{-1} Z' ) X \right) \right\} \\
		& {} \cdot \dpconc^{\ko-1} \frac{\Gamma(\dpconc + 1)}{\Gamma(\dpconc + N)} \prod_{k=1}^{\ko} (S_{N,k} - 1)!
\end{align*}

Now set $\theta := \exp(-\lambda^{2}/(2\sigma^{2}))$ and consider the limit $\sigma^{2} \rightarrow 0$. 
Then
\begin{align*}
	\lefteqn{ -\log \mbp(X,Z)
		=
		O(\log(\sigma^{2})) } \\
		& {} +  \frac{1}{2\sigma^{2}} \tr\left( X' (I_{N} - Z (Z'Z + \frac{\sigma^{2}}{\rho^{2}} I_{D})^{-1} Z' ) X \right) \\
		& {} + (\ko - 1) \frac{ \lambda^{2} }{ 2 \sigma^{2} }
			+ O(1)
\end{align*}

It follows that
\begin{align*}
	\lefteqn{ -2\sigma^{2} \log \mbp(X,Z) } \\
		&= \tr\left( X' (I_{N} - Z (Z'Z + \frac{\sigma^{2}}{\rho^{2}} I_{D})^{-1} Z' ) X \right) \\
		& {} + (\ko - 1) \lambda^{2} + O(\sigma^{2} \log(\sigma^{2}))
\end{align*}
We note that $\sigma^{2} \log(\sigma^{2}) \rightarrow 0$ as $\sigma^{2} \rightarrow 0$. Further note that $Z'Z$ is a diagonal $K \times K$ matrix with $(k,k)$ entry (call it $S_{N,k}$) equal to the number of indices in cluster $k$. $Z'Z$ is invertible since we assume no empty clusters are represented in $Z$. Then
\begin{align*}
	\lefteqn{ \lefteqn{ -2\sigma^{2} \log \mbp(X,Z) } } \\
		&\sim \tr\left( X' (I_{N} - Z (Z'Z)^{-1} Z' ) X \right) + (\ko - 1) \lambda^{2}
\end{align*}
as $\sigma^{2} \rightarrow 0$.

\subsection{More interpretable objective} \label{app:cdp_interp}

The objective for the collapsed Dirichlet process is more interpretable after some
algebraic manipulation. We describe here how the opaque $\tr\left( X' (I_{N} - Z (Z'Z)^{-1} Z' ) X \right)$
term can be written in a form more reminiscent of the $\sum_{k=1}^{\ko} \sum_{n: z_{n,k}=1} \| x_{n} - \mu_{k} \|^{2}$ term in the uncollapsed objective.
First, recall that $C := Z'Z$ is a $K \times K$ matrix with $C_{k,k} = S_{N,k}$ and $C_{j,k} = 0$ for $j \ne k$. Then
$C' := Z(Z'Z)^{-1}Z'$ is an $N \times N$ matrix with $C'_{n,m} = S_{N,k}^{-1}$ if and only if $z_{n,k} = z_{m,k} = 1$ and $C'_{n,m} = 0$ if $z_{n,k} \ne z_{m,k}$.
\begin{align*}
	\lefteqn{ \tr(X'(I_{N} - Z(Z'Z)^{-1} Z')X) } \\
		&= \tr(X'X) - \tr(X' Z (Z'Z)^{-1} Z'X) \\
		&= \tr(XX') - \sum_{d=1}^{D} \sum_{k=1}^{\ko} \sum_{n: z_{n,k} = 1} \sum_{m: z_{m,k} = 1}
			S_{N,k}^{-1} X_{n,d} X_{m,d} \\
		&= \sum_{k=1}^{\ko} \left[
			\sum_{n: z_{n,k} = 1} x_{n} x'_{n}
			- 2 S_{N,k}^{-1} \sum_{n: z_{n,k} = 1} x_{n} \sum_{m: z_{m,k} = 1} x'_{m}
			\right. \\
		& \left.
			{} + S_{N,k}^{-1} \sum_{n: z_{n,k} = 1} x_{n} \sum_{m: z_{m,k} = 1} x'_{m}
			\right] \\
		&= \sum_{k=1}^{\ko} \sum_{n: z_{n,k} = 1} \| x_{n} - S_{N,k}^{-1} \sum_{m: z_{m,k} = 1} x_{m,k} \|^{2} \\
		&= \sum_{k=1}^{\ko} \sum_{n: z_{n,k} = 1} \| x_{n} - \bar{x}^{(k)} \|^{2}
\end{align*}
for cluster-specific empirical mean $\bar{x}^{(k)} := S_{N,k}^{-1} \sum_{m: z_{m,k} = 1} x_{m,k}$ as in the main text.

\section{Collapsed BP-means objective derivation} \label{app:cbp_obj}

We continue to apply the usual recipe as in \app{dp_obj}. The generative model 
for collapsed BP-means is described in \mysec{collapsed}. The joint distribution
of the observed data $X$ and feature indicators $Z$ can be written as follows.
\begin{align*}
	\lefteqn{ \mbp(X,Z) = \mbp(X|Z) \mbp(Z) } \\
		&= 
		\left( (2\pi)^{ND/2} (\sigma^{2})^{(N-\ko )D/2} (\rho^{2})^{\ko D/2} |Z' Z + \frac{\sigma^{2}}{\rho^{2}} I_{D} |^{D/2} \right)^{-1} \\
		& {} \cdot \exp\left\{ - \frac{1}{2\sigma^{2}} \tr\left( X' (I_{N} - Z (Z'Z + \frac{\sigma^{2}}{\rho^{2}} I_{D})^{-1} Z' ) X \right) \right\} \\
		& {} \cdot \frac{\bpmass^{\ko}
			\exp\left\{ -\sum_{n=1}^{N} \frac{\bpmass}{n} \right\}
			}{ \prod_{h=1}^{H} \tk_{h}! }
			\prod_{k=1}^{\ko} \frac{(S_{N,k}-1)! (N - S_{N,k})!}{N!}
\end{align*}

Now set $\gamma := \exp(-\lambda^{2}/(2\sigma^{2}))$ and consider the limit $\sigma^{2} \rightarrow 0$.
Then
\begin{align*}
	\lefteqn{ -\log \mbp(X,Z) 
		= 
		O(\log(\sigma^{2})) } \\
		& {} +  \frac{1}{2\sigma^{2}} \tr\left( X' (I_{N} - Z (Z'Z + \frac{\sigma^{2}}{\rho^{2}} I_{D})^{-1} Z' ) X \right) \\
		& {} + \ko \frac{ \lambda^{2} }{ 2 \sigma^{2} }
			+ \exp(-\lambda^{2}/(2\sigma^{2})) \sum_{n=1}^{N} n^{-1}
			+ O(1)
\end{align*}

It follows that
\begin{align*}
	-2\sigma^{2} &\log \mbp(X,Z) 
		= \tr\left( X' (I_{N} - Z (Z'Z + \frac{\sigma^{2}}{\rho^{2}} I_{D})^{-1} Z' ) X \right) \\
		& {} + \ko \lambda^{2}
			+ O\left(\sigma^{2} \exp(-\lambda^{2}/(2\sigma^{2}))\right)
			+ O(\sigma^{2} \log(\sigma^{2})).
\end{align*}
But $\exp(-\lambda^{2}/(2\sigma^{2})) \rightarrow 0$ and $\sigma^{2} \log(\sigma^{2}) \rightarrow 0$ as $\sigma^{2} \rightarrow 0$. And $Z'Z$ is invertible so long as two features do not have identical membership (in which case we collect them into a single feature). So we have that $-2\sigma^{2} \log  \mbp(X,Z) \sim \tr\left( X' (I_{N} - Z (Z'Z)^{-1} Z' ) X \right) + \ko \lambda^{2}$.

\section{General multivariate Gaussian likelihood} \label{app:multivar_gauss}

Above, we assumed a multivariate spherical Gaussian likelihood for each cluster. This assumption
can be generalized in a number of ways. For instance, assume a general covariance matrix $\sigma^{2} \Sigma_{k}$ for positive scalar $\sigma^{2}$ and positive definite $D \times D$ matrix $\Sigma_{k}$. Then we assume the following likelihood model for data points assigned to the $k$th cluster ($z_{n,k} = 1$): $x_{n} \sim \gaus(\mu_{k}, \sigma^{2} \Sigma_{k})$. Moreover, assume an inverse Wishart prior on the positive definite matrix $\Sigma_{k}$: $\Sigma_{k} \sim W(\Phi^{-1}, \nu)$ for $\Phi$ a positive definite matrix and $\nu > D - 1$. Assume a prior $\mbp(\mu)$ on $\mu$ that puts strictly positive density on all real-valued $D$-length vectors $\mu$. For now we assume $K$ is fixed and that $\mbp(z)$ puts a prior that has strictly positive density on all valid clusterings of the data points. This analysis can be immediately extended to the varying cluster number case via the reasoning above. Then
\begin{align*}
	\lefteqn{ \mbp(x,z,\mu,\sigma^{2} \Sigma)  } \\ 
		&= \mbp(x|z,\mu,\sigma^{2} \Sigma) \mbp(z) \mbp(\mu) \mbp(\Sigma) \\
		&= 
		\prod_{k=1}^{\ko} \prod_{n: z_{n,k}=1} \gaus(x_{n} | \mu_{k}, \sigma^{2} \Sigma_{k}) \\
		& {} \cdot \mbp(z) \mbp(\mu)
		\cdot \prod_{k=1}^{K} \left[ \frac{| \Phi |^{\nu/2}}{2^{\nu D/2} \Gamma_{D}(\nu/2)} | \Sigma_{k}|^{-\frac{\nu+D+1}{2}} \right. \\
			& \left. {} \cdot \exp\left\{-\frac{1}{2} \tr( \Phi \Sigma_{k}^{-1} ) \right\} \right],
\end{align*}
where $\Gamma_{D}$ is the multivariate gamma function.
Consider the limit $\sigma^{2} \rightarrow 0$. Set $\nu = \lambda^{2}/\sigma^{2}$
for some constant $\lambda^{2}: \lambda^{2} > 0$. Then
\begin{align*}
	\lefteqn{ - \log \mbp(x,z,\mu,\sigma^{2} \Sigma)  } \\ 
		&= 
		\sum_{k=1}^{K} \sum_{n: z_{n,k}=1} \left[ 
			O(\log \sigma^{2})
			+ \frac{1}{2 \sigma^{2}} (x_{n} - \mu_{k})' \Sigma_{k}^{-1} (x_{n} - \mu_{k})
			\right] \\
		& {} + O(1)
		+ \sum_{k=1}^{K} \left[ -\frac{1}{2\sigma^{2}} \lambda^{2} \log | \Phi | + \frac{D}{2 \sigma^{2}} \lambda^{2} \log 2 \right. \\
		& \left. {} + \log \Gamma_{D}(\lambda^{2}/(2\sigma^{2})) + \left( \frac{\lambda^{2}}{2\sigma^{2}} + \frac{D+1}{2} \right) \log | \Sigma_{k} | + O(1) \right]
\end{align*}

So we find
\begin{align*}
	\lefteqn{ - 2 \sigma^{2} \left[ \log \mbp(x,z,\mu,\sigma^{2} \Sigma) + \log \Gamma_{D}(\lambda^{2}/(2\sigma^{2})) \right]} \\ 
		&\sim \sum_{k=1}^{K} \sum_{n: z_{n,k}=1} (x_{n} - \mu_{k})'
			\Sigma_{k}^{-1} (x_{n} - \mu_{k}) \\
		& {} + \sum_{k=1}^{K} \lambda^{2} \log |\Sigma_{k}|
			+ O(\sigma^{2}).
\end{align*}
Letting $\sigma^{2} \rightarrow 0$, the righthand side becomes
$$
	\sum_{k=1}^{K} \sum_{n: z_{n,k}=1} (x_{n} - \mu_{k})'
\Sigma_{k}^{-1} (x_{n} - \mu_{k})
+ \sum_{k=1}^{K} \lambda^{2} \log |\Sigma_{k}|,
$$
the final form of the objective.

If the $\Sigma_{k}$ are known, they may be inputted and the objective may be
optimized over the cluster means and cluster assignments. In general, though, the
resulting optimization problem is
$$
	\min_{z,\mu,\Sigma} \sum_{k=1}^{K} \left[ \sum_{n: z_{n,k}=1} (x_{n} - \mu_{k})'
			\Sigma_{k}^{-1} (x_{n} - \mu_{k})
			+ \lambda^{2} \log |\Sigma_{k}| \right]
$$
That is, the squared Euclidean distance in the classic K-means objective
function has been replaced with a Mahalanobis distance, and we have
added a penalty term on the size of the $\Sigma_{k}$ matrices (with $\lambda^{2}$
modulating the penalty as in previous examples). This objective
is reminiscent of that proposed by \citet{sung:1998:example}.

\section{Proof of BP-means local convergence} \label{app:bp_means_conv}

The proof of \prop{ibp_means} is as follows.

\begin{proof}
By construction, the first step in any iteration does not increase the objective. The second step starts by deleting any features that have the same index collection as an existing feature. Suppose there are $m$ such features with indices $J$ and we keep feature $k$. By setting $A_{k,\cdot} \leftarrow \sum_{j \in J} A_{j,\cdot}$, the objective is unchanged. Next, note
\begin{align}
	\label{eq:grad_A}
	\nabla_{A} \tr[(X - ZA)' (X - ZA)]
		&= 2 Z' (X - ZA).
\end{align}
Setting the gradient to zero, we find that $A = (Z' Z)^{-1} Z' X$ solves the equation for $A$
and therefore minimizes the objective with respect to $A$ when $Z' Z$ is invertible, as we have already guaranteed.

Finally, since there is only a finite number of feature allocations in which each data point has at most one feature unique to only that data point and no features containing identical indices (any extra such features would only increase the objective due to the penalty), the algorithm cannot visit more than this many configurations and must finish in a finite number of iterations.
\end{proof}

\end{document}